%% file: arxiv.tex
\documentclass[runningheads]{llncs}

\usepackage{eccv}

\usepackage{eccvabbrv}

\usepackage{graphicx}
\usepackage{booktabs}
\usepackage{multirow}
\usepackage{microtype}
\newcommand{\tworow}[2]{\begin{tabular}[c]{@{}c@{}}#1\vspace{-2pt}\\#2\end{tabular}}
\usepackage[accsupp]{axessibility}  

\makeatletter
\def\@fnsymbol#1{%
  \ensuremath{%
    \ifcase#1
    \or *%
    \or \dagger%
    \or \ddagger%
    \or \mathsection%
    \or \mathparagraph%
    \else \@ctrerr
    \fi
  }%
}
\makeatother

\usepackage{hyperref}

\usepackage{orcidlink}

\begin{document}

\title{TriMotion: Modality-Agnostic Camera Control \newline for Video Generation} 



\author{Seunghyun Shin\inst{1,2}$^{*}$\orcidlink{0009-0006-3012-9675} \and
Jifei Song\inst{2}\orcidlink{0000-0002-3381-6685} \and
Wooseok Jeon\inst{3}\orcidlink{0009-0003-4511-8168}\and
\newline Hae-Gon Jeon\inst{3}$^{\dagger}$\orcidlink{0000-0003-1105-1666} \and
Jiankang Deng\inst{4}$^{\dagger}$\orcidlink{0000-0002-3709-6216}
}

\authorrunning{Shin et al.}

\institute{$^{1}$ GIST \quad $^{2}$ Huawei Noah's Ark Lab \quad $^{3}$ Yonsei University \quad $^{4}$ Imperial College London \\
Project page: \url{https://seunghyuns98.github.io/TriMotion/}}



\begingroup
\renewcommand{\thefootnote}{\fnsymbol{footnote}}
\footnotetext[1]{Work done during an internship at Huawei Noah's Ark Lab.}
\footnotetext[2]{Corresponding authors.}
\endgroup

\maketitle

\begin{figure}[h]
\vspace{-8mm}
\centering
\includegraphics[width=0.96\columnwidth]{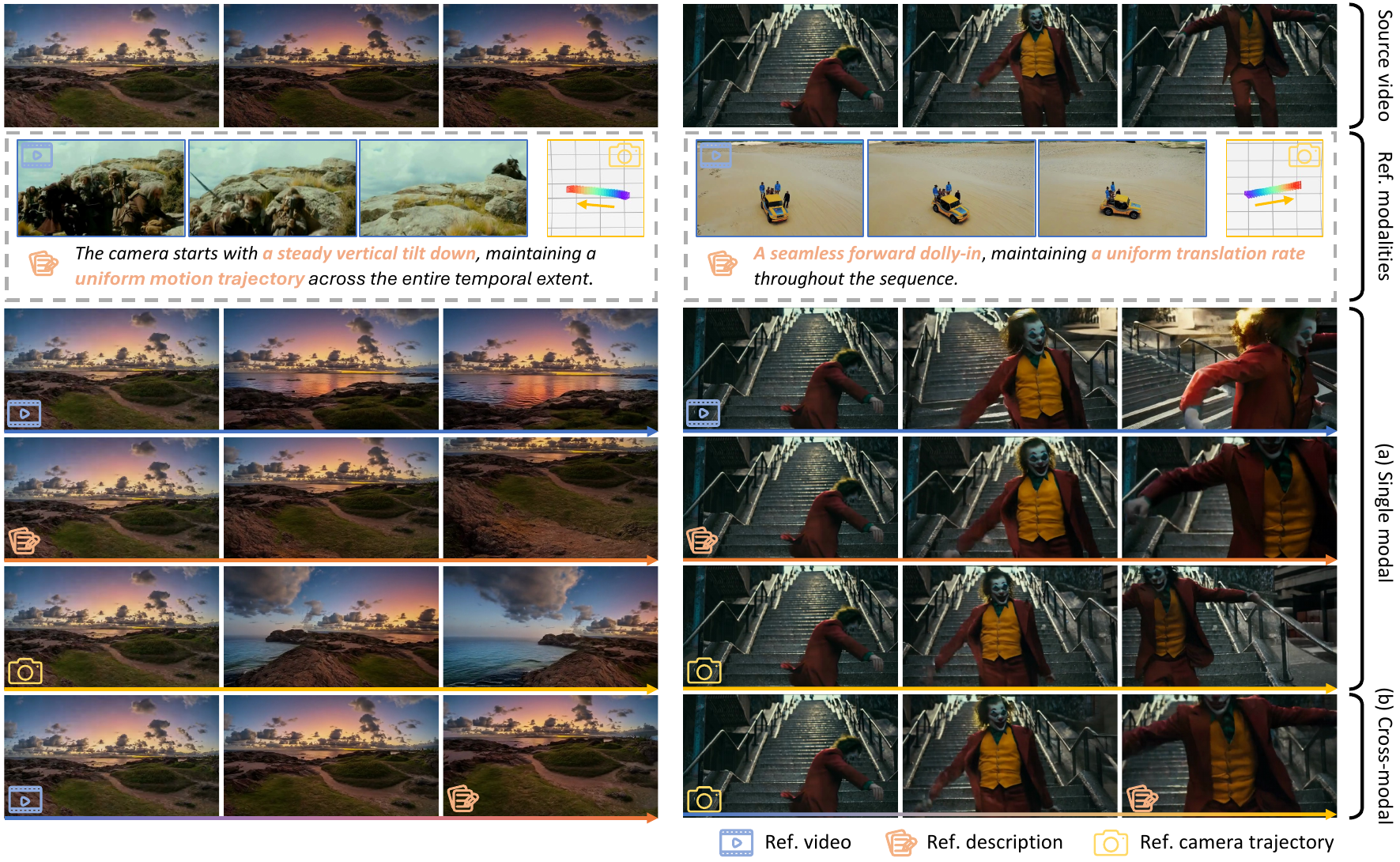}
\vspace{-3mm}
\caption{\textbf{Camera-controlled video generation results of TriMotion.}
(a) TriMotion enables precise camera control from three reference modalities. (b) It also supports cross-modal camera control, allowing camera motion to be transferred across modalities.}
\label{fig:teaser}
\vspace{-12mm}
\end{figure}

\begin{abstract}
Camera motion control is essential for directing viewpoint changes in generative systems. 
However, existing methods typically condition the generation process on a single specific modality, such as explicit pose trajectories or reference videos, limiting their ability to support heterogeneous user inputs. 
To address this limitation, we present \textit{TriMotion}, a modality-agnostic framework for camera-controlled video generation that maps video, pose, and text inputs, describing the same camera trajectory into a shared motion embedding space. 
Learning such a space requires synchronized supervision across modalities.
Therefore, we build the \textit{Motion Triplet Dataset} by extending a Multi-Cam Video Dataset with geometry-grounded motion descriptions derived from camera extrinsics.
We further introduce a latent motion consistency objective that leverages the motion embedding space to encourage the generated video to follow the target camera trajectory directly in latent space, avoiding the cost of pixel-space decoding.
Extensive experiments show that TriMotion generates high-quality videos that accurately follow the target camera trajectories across all three modalities. 
Beyond standard generation, the shared motion embedding space also enables flexible applications such as sequential motion composition and cross-modal motion interpolation.
\vspace{-3mm}
\end{abstract}

\section{Introduction}
\label{intro}
Camera movement has long been regarded as a fundamental technique for organizing cinematic space and directing viewer attention~\cite{bordwell2008film}. 
By modulating viewpoint over time, filmmakers shape atmosphere and emotional emphasis within a scene. 
For example, in the film \textit{Vertigo}, the dolly zoom distorts spatial perception to externalize the protagonist’s psychological instability, a technique frequently analyzed in film studies for its affective and perceptual impact~\cite{wood2002hitchcock}.
Beyond traditional cinema, controlling camera motion has become critical in modern generative systems, where viewpoint dynamics strongly influence realism, narrative coherence, and user intent realization.

Driven by the rapid advancement of conditional video diffusion models~\cite{hong2022cogvideo, blattmann2023stable, blattmann2023align ,bar2024lumiere, xing2024dynamicrafter, yang2024cogvideox, wan2025, jeon2026motion}, recent approaches have actively explored various conditioning signals to direct camera movement. 
However, existing camera-control methods are often restricted to a single specific input reference modality.
Pose-conditioned methods~\cite{wang2024motionctrl, he2024cameractrl, he2025cameractrl, zheng2024cami2v, bai2025recammaster, bahmani2024vd3d, bahmani2025ac3d, yu2025trajectorycrafter, gu2025diffusion, vanhoorick2024gcd, zhang2025recapture, seo2025vidcamedt, jeong2025reangle} require explicit camera trajectories in the geometric manner which is not intuitive for general users and difficult to specify accurately. 
Reference-video-based approaches~\cite{luo2025camclonemaster, hu2024motionmaster} implicitly encode motion patterns, yet they lack explicit trajectory control and offer limited flexibility. 
Text descriptions, while intuitive, remain underexplored for camera control due to their lack of explicit temporal structures and geometric constraints, which makes it difficult to ensure precise and physically consistent trajectory execution.
Although each modality offers advantages in terms of accessibility and geometric precision, existing methods typically restrict users to a single modality, limiting the flexibility of generative systems. 
This motivates a shared motion representation that bridges heterogeneous inputs and enables consistent camera motion control across modalities.

To this end, we propose TriMotion, a unified framework for camera-controlled video generation. 
Trimotion maps diverse control signals including video, pose, and text inputs, into a single continuous motion representation.
By aligning equivalent camera trajectories across modalities, we preserve both temporal dynamics and physical consistency.
This shared representation enables consistent camera controls in latent video diffusion models regardless of the input modality.

Since training such a cross-modal space inherently demands synchronized, multi-modal supervision, we curate the Motion Triplet Dataset. 
Leveraging a Large Language Model~(LLM)-based pipeline, we translate explicit camera extrinsics from the Multi-Cam Video Dataset~\cite{bai2025recammaster} into geometry-grounded motion descriptions. 
This process pairs multi-view video data with geometry-grounded text, providing the essential foundation for aligning divergent modalities without sacrificing geometric fidelity.

Built on the Motion Triplet Dataset, TriMotion consists of two main components: 
First, we learn a unified motion representation using global contrastive alignment, temporal token matching, and pose regression, so that the representation captures both trajectory semantics and camera geometry. 
Second, we introduce a latent motion consistency objective implemented through a motion embedding predictor, which regularizes the diffusion backbone directly in latent space without repeated pixel-space decoding. 
A dedicated motion embedding predictor ensures geometric alignment entirely in the latent domain, thereby improving controllability while bypassing traditional decoding bottlenecks.

Empirical evaluations confirm that TriMotion not only establishes new state-of-the-art benchmarks for camera-controllable video generation but also pioneers novel interaction paradigms, successfully enabling sequential motion composition and cross-modal motion interpolation. (See Fig. \ref{fig:teaser}).


\section{Related Work}
\label{related_works}

\noindent\textbf{Camera-Controllable Video Generation}
Camera-controllable video generation aims to follow a user-specified camera trajectory while preserving temporal and geometric consistency. 
Existing methods can be broadly grouped into three families.
First, pose-conditioned methods condition pretrained video generators~\cite{wan2025, chen2024videocrafter2, hong2022cogvideo, yang2024cogvideox, blattmann2023stable} on explicit camera trajectories.
For example, MotionCtrl~\cite{wang2024motionctrl} directly incorporates camera extrinsics into temporal modules, while methods such as CameraCtrl~\cite{he2025cameractrl}, VD3D~\cite{bahmani2024vd3d}, AC3D~\cite{bahmani2025ac3d}, and I2VControl-Camera~\cite{feng2024i2vcontrol} adopt advanced geometric parameterizations like Pl\"ucker rays or point trajectories to improve control.
Second, geometry-aware image-to-video methods enhance cross-view consistency through 3D reasoning.
CamCo~\cite{xu2024camco} and CamI2V~\cite{zheng2024cami2v} incorporate epipolar-constrained attention, whereas RealCam-I2V~\cite{li2025realcam} further leverages metric-depth-based scene reconstruction and an interactive 3D interface to design camera trajectories on real images.
Finally, reference-based methods avoid explicit camera parameters by transferring motion from exemplar videos. 
MotionMaster~\cite{hu2024motionmaster} and MotionClone~\cite{ling2024motionclone} extract motion cues from temporal attention maps to guide generation.

\noindent\textbf{Camera-controlled Video-to-Video Generation}
Camera-controlled video-to-video generation aims to re-render an input clip under a new viewpoint while preserving scene identity, appearance, and dynamics.
Existing methods can be categorized into two directions. 
The first direction uses conditioning-based control, where generative models are guided by target camera parameters or motion cues.
GCD~\cite{vanhoorick2024gcd} introduces synthetic multi-view training, while ReCamMaster~\cite{bai2025recammaster} improves robustness on real videos through stronger video conditioning and a large synchronized multi-camera dataset. 
DaS~\cite{gu2025diffusion} uses 3D tracking videos as control signals, GS-DiT~\cite{bian2025gs} builds pseudo-4D Gaussian fields from dense 3D point tracking. 
Trajectory Attention~\cite{xiao2024trajectory} injects pixel trajectories through a dedicated attention branch. 
CamCloneMaster~\cite{luo2025camclonemaster} transfers camera motion from a reference video through token concatenation, supporting both Image-to-Video~(I2V) and Video-to-Video~(V2V) without camera parameters. 
Another approach follows warping-based view transformation, where source content is first projected or warped into target viewpoints using estimated geometry and then refined through a generative model. 
ReCapture~\cite{zhang2025recapture} first produces anchor videos and refines them with masked video fine-tuning, while TrajectoryCrafter~\cite{yu2025trajectorycrafter}, Vid-CamEdit~\cite{seo2025vidcamedt}, and Reangle-A-Video~\cite{jeong2025reangle} use point clouds or estimated geometry to guide diffusion-based completion under novel views.

\noindent\textbf{TriMotion's Approach}
Unlike task-specific or modality-restricted approaches, TriMotion learns a shared motion embedding space across video, pose, and text inputs. This unified approach is enabled by the \textit{Motion Triplet Dataset}, which provides synchronized supervision across the three modalities, enabling flexible and consistent camera control across different input types.

\section{Motion Triplet Dataset}
\label{sec:dataset}

The Motion Triplet Dataset is built upon the Multi-Cam Video Dataset~\cite{bai2025recammaster} which contains 136K high-fidelity Unreal Engine 5~\cite{epicgames_unrealengine5} videos spanning 13.6K dynamic scenes across 40 diverse 3D environments, with 122K unique camera trajectories and accurate ground-truth camera extrinsics. 
While this dataset provides synchronized videos and poses, it lacks textual descriptions of camera motion. 
Accordingly, we design an automated geometry-grounded captioning pipeline that converts raw camera trajectories into natural language.

Directly feeding raw camera pose sequences to a Large Language Model (LLM) is often suboptimal, since pose trajectories are continuous numerical signals with temporal dependencies, whereas LLMs are optimized for discrete language tokens. 
Recent works highlight modality gaps when applying LLMs to numerical time-series data and show that reliable temporal numerical reasoning remains challenging~\cite{jin2023time,arai2025evaluating}.
To mitigate this, we first convert them into relative trajectories anchored at the first frame, and compute frame-wise translation and rotation changes given per-frame camera extrinsics. 
Small variations below fixed thresholds are treated as stationary, while significant motions are mapped to canonical camera operations such as Dolly, Pan, and Tilt, then organized into temporally ordered motion phases. 
This produces a compact symbolic sequence that preserves the essential structure of the trajectory.

We use these symbolic sequences to prompt Qwen3-4B-Instruct~\cite{yang2025qwen3} to generate two levels of motion descriptions: a short summary of the overall trajectory and a detailed paragraph describing temporal transitions and speed changes. 
The model is instructed to distinguish simultaneous from sequential motion and to avoid explicit metric values, matching the qualitative style of realistic user prompts. 
The resulting Motion Triplet Dataset provides aligned supervision over videos, camera trajectories, and motion descriptions for learning a unified motion representation. 
We further validate the quality of the generated descriptions through a user study in Sec.~\ref{sec:user_study}. Further details are provided in Sec.~\ref{sec:supp_dataset}.

\begin{figure}[t]
\centering
\includegraphics[width=\columnwidth]{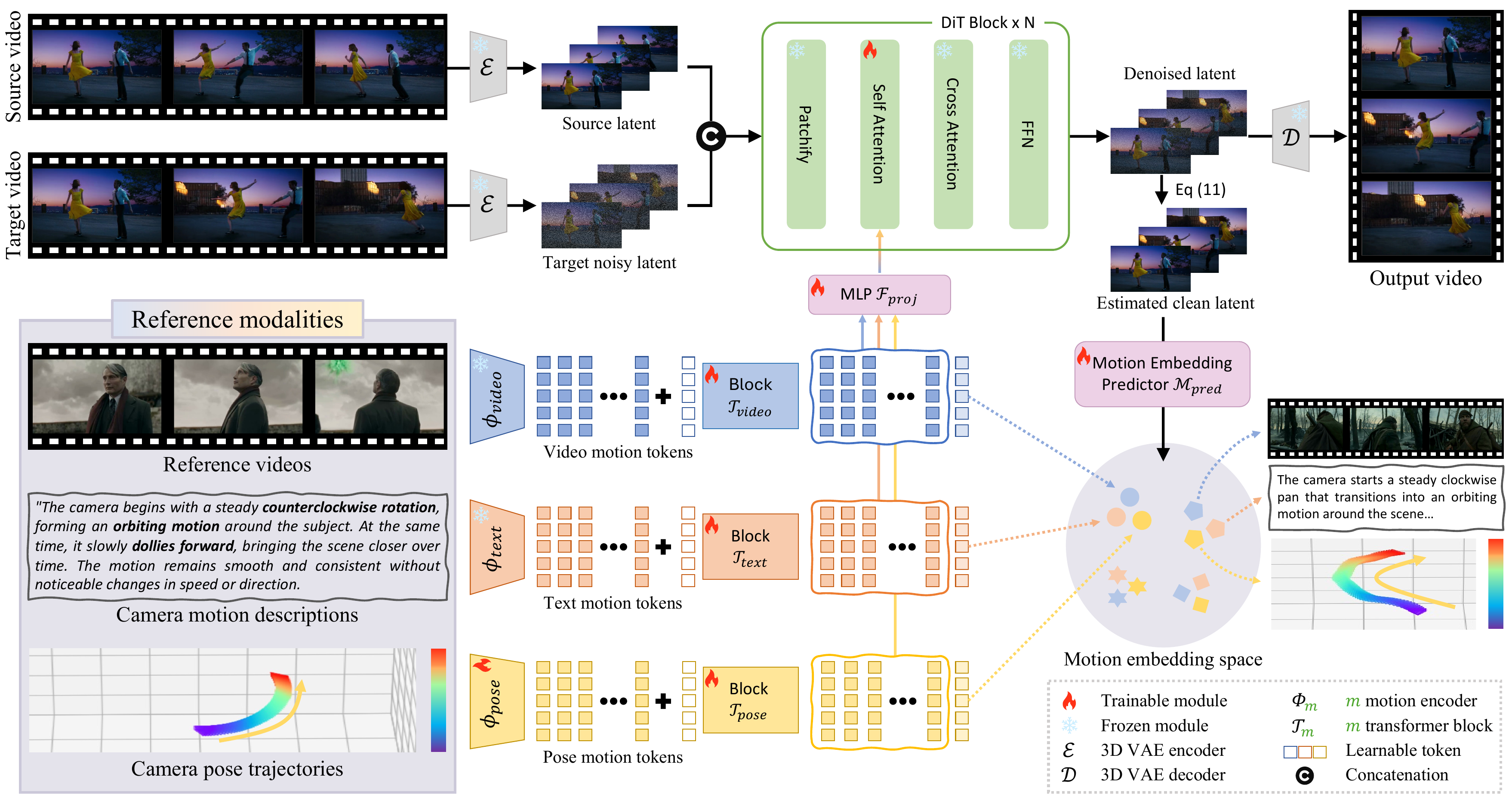}
\caption{\textbf{Overview of TriMotion.} TriMotion maps video, pose, and text motion inputs into a unified motion embedding space and uses the resulting embedding to condition the latent video diffusion backbone for camera-controlled video generation.}
\label{fig:overview}
\vspace{-5mm}
\end{figure}

\section{Method}
\label{sec:method}

In this section, we describe TriMotion, a framework for cross-modal camera motion control built on the latent video diffusion transformer~\cite{wan2025}.
We first introduce the diffusion backbone and motion conditioning setup (Sec.~\ref{sec:preliminary}). 
We then describe two key components of the framework: a Unified Motion Embedding Space that aligns video, text, and pose inputs in a shared representation (Sec.~\ref{sec:unified_space}), and a Latent Motion Consistency constraint that preserves the target camera trajectory during generation (Sec.~\ref{sec:latent}). 
Finally, we present the overall training objective and strategy (Sec.~\ref{sec:overall_objective}). 
An overview is shown in Fig.~\ref{fig:overview}.

\subsection{Diffusion Backbone and Motion Conditioning}
\label{sec:preliminary}

Given a source video $x_{\mathrm{src}} \in \mathbb{R}^{F \times C \times H \times W}$ and a target video $x_{\mathrm{tgt}} \in \mathbb{R}^{F \times C \times H \times W}$, a 3D Variational Autoencoder~(VAE) encoder $\mathcal{E}$ compresses them into latent representations
\begin{equation}
\mathbf{z}_{\mathrm{src}} = \mathcal{E}(x_{\mathrm{src}}), \qquad
\mathbf{z}_{0} = \mathcal{E}(x_{\mathrm{tgt}}),
\end{equation}
where the temporal and spatial dimensions are downsampled by factors of 4 and 8, respectively. 
A decoder $\mathcal{D}$ reconstructs the video such that $x \approx \mathcal{D}(\mathbf{z})$.

Following rectified flow training~\cite{esser2024scaling, liu2022flow, lipman2022flow}, we sample $\epsilon \sim \mathcal{N}(0, I)$ and define a noisy target latent as:
\begin{equation}
\mathbf{z}_{t} = (1 - t)\mathbf{z}_{0} + t\epsilon, \quad t \in [0,1].
\end{equation}
The diffusion input is then formed by concatenating the clean source latent and the noisy target latent along the temporal dimension: $ \tilde{\mathbf{z}}_{t} = [\mathbf{z}_{\text{src}} ; \mathbf{z}_{t}]$.

In addition to the latent input, the denoiser is conditioned on a global appearance description $y$, the first source frame $I$, and a target motion embedding sequence $\mathbf{e}_m$ from the unified motion space described in Sec.~\ref{sec:unified_space}. 
The description and first frame are encoded by a frozen T5 encoder~\cite{raffel2020exploring} and a frozen CLIP image encoder~\cite{radford2021learning}, respectively. 
To incorporate target motion, we inject $\mathbf{e}_m$ into each transformer block through a block-specific projection MLP $\mathcal{F}_{\mathrm{proj}}$ and residual addition:
\begin{equation}
\mathbf{h}_{\mathrm{in}} = \mathbf{h} + \mathcal{F}_{\mathrm{proj}}(\mathbf{e}_m),
\end{equation}
where $\mathcal{F}_{\mathrm{proj}}$ maps the motion embedding to the hidden feature layout of the diffusion backbone. 
This residual motion injection steers the temporal dynamics of the denoising process while preserving pretrained spatial priors.

The denoiser $\mathbf{v}_{\theta}$ is optimized to predict the target velocity
$\mathbf{u}_{t}(\mathbf{z}_{0}, \epsilon) = \epsilon - \mathbf{z}_{0}$:
\begin{equation}
\mathcal{L}_{\mathrm{denoise}} =
\mathbb{E}_{t,\mathbf{z}_{\mathrm{src}},\mathbf{z}_{0},\epsilon,y,I,\mathbf{e}_m}
\left[
\left\|
\mathbf{v}_{\theta}(\tilde{\mathbf{z}}_{t}, t, y, I, \mathbf{e}_m) - \mathbf{u}_{t}(\mathbf{z}_{0}, \epsilon)
\right\|_2^2
\right].
\end{equation}
This design allows the model to preserve scene appearance from the source video while generating the target video under the desired camera motion.
To support both I2V and V2V within a unified formulation, we construct the source latent as a pseudo-video latent whose first frame is obtained by encoding the input image, while the remaining latent frames are zero-filled for I2V.

\subsection{Unified Motion Embedding Space}
\label{sec:unified_space}

To enable modality-agnostic camera control, we map video, text, and pose inputs into a shared motion embedding space that captures both global trajectory intent and temporal camera dynamics. 
Given an input from modality $m \in \{\mathrm{video}, \mathrm{text}, \mathrm{pose}\}$, a modality-specific encoder $\Phi_m$ produces a sequence of $N$ motion tokens 
$\mathbf{h}_m \in \mathbb{R}^{N \times D}$, where $N$ is a fixed token length shared across modalities and $D$ denotes the embedding dimension. 
We prepend a learnable \textit{global motion token} $\mathbf{h}_g \in \mathbb{R}^{1 \times D}$ to aggregate trajectory-level information. 
The concatenated sequence is processed by a lightweight temporal Transformer $\mathcal{T}_m$:
\begin{equation}
\mathbf{e}_m = \mathcal{T}_m([\mathbf{h}_g ; \mathbf{h}_m]), \quad \mathbf{e}_m \in \mathbb{R}^{(N+1) \times D}
\end{equation}
The first output token $\mathbf{e}_m^{(0)}$ encodes the global motion intent, while the remaining tokens $\mathbf{e}_m^{(1:N)}$ represent temporally ordered camera dynamics.

\vspace{-1.1mm}
\paragraph{Video Motion Encoder~$\Phi_{\mathrm{video}}$.}
To extract camera motion cues from video, we adopt the feature aggregation module of VGGT~\cite{wang2025vggt}. 
The encoder patchifies input frames and appends a learnable camera token to each frame sequence. 
Through Alternating-Attention blocks that interleave frame-wise and global self-attention, multi-view 3D geometric information is aggregated into these camera tokens.
We use the resulting camera tokens as the motion token sequence $\mathbf{h}_{\mathrm{video}}$.

\vspace{-1.1mm}
\paragraph{Text Motion Encoder~$\Phi_{\mathrm{text}}$.}
Unlike camera motion, which unfolds over time, text provides a static description without explicit temporal structure. 
We encode the sentence using a frozen T5 encoder~\cite{raffel2020exploring} to obtain contextualized text tokens. 
To lift this representation into a temporal motion sequence, we introduce $N$ learnable motion queries that cross-attend to the text tokens via multi-head attention. 
The resulting query features form the motion token sequence $\mathbf{h}_{\mathrm{text}}$, which represents temporal camera dynamics inferred from language.

\vspace{-1.1mm}
\paragraph{Pose Motion Encoder~$\Phi_{\mathrm{pose}}$.}
We represent the camera pose at each frame as a flattened $3\times4$ camera extrinsic matrix.
To project these geometrically explicit parameters into the shared $D$-dimensional embedding space, we apply a frame-wise multi-layer perceptron (MLP) with GELU activations. 
This yields the motion token sequence $\mathbf{h}_{\mathrm{pose}}$ while preserving the original trajectory structure.

\vspace{-1.1mm}
\paragraph{}To align equivalent camera trajectories across modalities, we train the motion encoders with a composite objective consisting of global alignment, temporal synchronization, and geometric fidelity regularization.

\noindent\textbf{Global Alignment.}
We align the global motion tokens $\mathbf{e}_m^{(0)}$ across modalities using an InfoNCE objective computed over modality pairs:
\begin{equation}
\mathcal{L}_{\mathrm{NCE}}
= \sum_{(u,v)\in\mathcal{P}}
\frac{1}{B}\sum_{i=1}^{B}
-\log
\frac{
\exp\left(\mathrm{sim}(\mathbf{u}^{(0)}_i,\mathbf{v}^{(0)}_i)/\tau\right)
}{
\sum_{j=1}^{B}\exp\left(\mathrm{sim}(\mathbf{u}^{(0)}_i,\mathbf{v}^{(0)}_j)/\tau\right)
},
\end{equation}
where $\mathcal{P}=\{(\mathbf{e}_{\mathrm{video}},\mathbf{e}_{\mathrm{text}}),(\mathbf{e}_{\mathrm{video}},\mathbf{e}_{\mathrm{pose}}),(\mathbf{e}_{\mathrm{text}},\mathbf{e}_{\mathrm{pose}})\}$,
$B$ is the batch size, $\tau$ is a temperature parameter, and $\mathrm{sim}(\cdot,\cdot)$ denotes cosine similarity.
In practice, we symmetrize the loss by swapping the anchor and target modalities.

\noindent\textbf{Temporal Synchronization.} 
To enforce fine-grained temporal alignment, we minimize the cosine distance between corresponding temporal tokens:
\begin{equation}
\mathcal{L}_{\mathrm{temp}} = \sum_{(u,v)\in\mathcal{P}} \frac{1}{B\cdot N} \sum_{i=1}^{B} \sum_{k=1}^{N} \left(1-\mathrm{sim}(\mathbf{u}^{(k)}_i, \mathbf{v}^{(k)}_i)\right).
\end{equation}

\noindent\textbf{Geometric fidelity regularization.} 
Contrastive objectives are primarily driven by relative similarity on normalized embeddings, which may weaken absolute geometric cues in the learned space~\cite{gupta2023structuring}.
To retain physically meaningful camera motion, we attach a shared pose regressor that predicts camera extrinsics from each modality embedding:
\begin{equation}
\mathcal{L}_{\mathrm{pose}} = \sum_{m \in \{\mathrm{video, text, pose}\}} \frac{1}{B} \sum_{i=1}^{B} \|\hat{\mathbf{p}}_{m,i} - \mathbf{p}_{\mathrm{GT},i}\|_1,
\end{equation}
where $\hat{\mathbf{p}}_{m,i}$ denotes the pose predicted from modality $m$ for the $i$-th sample, and $\mathbf{p}_{\mathrm{GT},i}$ is the corresponding ground-truth camera pose. 
This regularization encourages the motion embeddings to preserve consistent 3D camera geometry.

The overall objective for learning the unified motion space is
\begin{equation}
\mathcal{L}_{\mathrm{align}}
=
\mathcal{L}_{\mathrm{NCE}}
+
\lambda_t \mathcal{L}_{\mathrm{temp}}
+
\lambda_p \mathcal{L}_{\mathrm{pose}},
\end{equation}
where $\lambda_t$ and $\lambda_p$ balance the different loss terms.

\subsection{Latent Motion Consistency}
\label{sec:latent}
Although the target motion embedding guides generation, it does not by itself guarantee strict geometric consistency in the generated trajectory. 
To improve trajectory fidelity, we introduce a Motion Embedding Predictor $\mathcal{M}_{\mathrm{pred}}$, which consists of 3D convolutions and a temporal Transformer encoder. 
Given a clean latent $\mathbf{z}_0$, the predictor estimates a motion embedding sequence: $\hat{\mathbf{e}} = \mathcal{M}_{\mathrm{pred}}(\mathbf{z}_0).$

We first train the predictor using a dual-granularity cosine similarity loss:
\begin{equation}
\mathcal{L}_{\mathrm{motion}} = \left(1 - \mathrm{sim}(\hat{\mathbf{e}}^{(0)}, \mathbf{e}_{\mathrm{video}}^{(0)})\right) +
\lambda_h \frac{1}{N}\sum_{k=1}^{N}\left(1 - \mathrm{sim}(\hat{\mathbf{e}}^{(k)}, \mathbf{e}_{\mathrm{video}}^{(k)})\right),
\end{equation}
where $\lambda_h$ balances global and token-wise alignment. 
The first term aligns the global motion intent, while the second enforces frame-wise temporal consistency.

Since the diffusion backbone may yield imperfect estimates of the clean latent, we improve robustness by injecting mild diffusion noise during predictor training, sampling from lower diffusion timesteps. 
After convergence, $\mathcal{M}_{\mathrm{pred}}$ is frozen and used during diffusion backbone training.

During diffusion training, we reconstruct the clean target latent $\hat{\mathbf{z}}_0 $  from the noisy target latent using the predicted velocity $\mathbf{v}_{\theta}(\tilde{\mathbf{z}}_{t}, t, y, I, e_{m})$:
\begin{equation}
\hat{\mathbf{z}}_0
=
\mathbf{z}_t
-
t\,\mathbf{v}_{\theta}(\tilde{\mathbf{z}}_{t}, t, y, I, \mathbf{e}_m).
\end{equation}
We then feed $\hat{\mathbf{z}}_0$ into the frozen predictor and apply the same motion consistency loss with respect to the target motion embedding $\mathbf{e}_m$. 
This latent-space constraint explicitly encourages the generated trajectory to match the target motion without requiring pixel-space reconstruction or decoding of high-resolution video frames.

\subsection{Training Objective and Strategy}
\label{sec:overall_objective}

We train our generative backbone by jointly optimizing the denoising loss and the latent motion consistency loss. 
Given the motion embedding predicted from the reconstructed clean latent, 
$\hat{\mathbf{e}}_{\mathrm{gen}} = \mathcal{M}_{\mathrm{pred}}(\hat{\mathbf{z}}_0)$, 
we enforce alignment with the target motion embedding $\mathbf{e}_m$ using $\mathcal{L}_{\mathrm{motion}}$. 
The overall objective is defined as:
\begin{equation}
    \mathcal{L}_{\mathrm{total}}
    = \mathcal{L}_{\mathrm{denoise}}
            + \lambda_{m}\mathcal{L}_{\mathrm{motion}}(\hat{\mathbf{e}}_{\mathrm{gen}}, \mathbf{e}_m),
\end{equation}
where $\lambda_{m}$ controls a trade-off between generative fidelity and motion consistency.

Following the efficient fine-tuning paradigm of CamCloneMaster~\cite{luo2025camclonemaster}, we update only the 3D spatial-temporal attention layers and the block-specific projection MLPs $\mathcal{F}_{\mathrm{proj}}$ within the diffusion backbone. 
This preserves the pretrained generative prior while efficiently adapting the model to motion-conditioned video generation. 
To support both I2V and V2V within a single framework, we use balanced joint training. At each iteration, the model receives either a full source sequence (V2V) or a single-frame source condition (I2V) with equal probability. 
This strategy encourages robust performance across both tasks without bias toward a particular source format.

\section{Experiments}
\label{sec:experiment}


\subsection{Experimental Setup}
\label{sec:exp_setup}

\noindent\textbf{Implementation Details.}
We train our method on the Motion Triplet Dataset described in Sec.~\ref{sec:dataset}, with 1\% of the data held out for validation. 
The unified motion space is trained for 100 epochs, the Motion Embedding Predictor for 10 epochs, and the diffusion backbone for 10K iterations. 
All models are trained on 4 NVIDIA H200 GPUs using the AdamW optimizer with $\beta_1 = 0.9$, $\beta_2 = 0.999$, weight decay $0.01$, and a learning rate of $1\times10^{-4}$.

\noindent\textbf{Evaluation Set.}
Our evaluation set consists of 500 source videos from Koala-36M~\cite{wang2025koala} and 500 video-pose pairs from RealEstate10K~\cite{zhou2018stereo}. 
Koala-36M provides diverse real-world videos with scene-level text descriptions, while RealEstate10K provides videos with corresponding per-frame camera poses. 
For samples with missing motion descriptions, we generate them from pose trajectories following the pipeline in Sec.~\ref{sec:dataset}, enabling evaluation across all three input modalities.

\noindent\textbf{Evaluation Metrics.}
We evaluate performance from three perspectives: visual quality, dynamic quality, and camera motion accuracy. 
For visual quality, we report FVD~\cite{unterthiner2018towards}, FID~\cite{heusel2017gans}, CLIP Score, and the Image Quality and Aesthetic Quality metrics from VBench~\cite{huang2024vbench}, which assess visual clarity and perceptual appeal. 
For dynamic quality, we report CLIP-F~\cite{bai2025recammaster} together with VBench metrics including Motion Smoothness, Temporal Flickering, and Subject/Background Consistency. 
For camera motion accuracy, we estimate camera poses using MegaSaM~\cite{li2025megasam} and compute Translation Error ($E_{\mathrm{trans}}$), Rotation Error ($E_{\mathrm{rot}}$), and Camera Motion Consistency (CamMC)~\cite{zheng2024cami2v}. 
For V2V, we further report FVD-V~\cite{xie2024sv4d} and CLIP-V to measure temporal quality and source-content preservation.

\input{Tables/quantitative}

\subsection{Comparison with State-of-the-Art Methods}
\label{sec:comparison}

\noindent\textbf{Baseline Models.}
We compare TriMotion with state-of-the-art camera-controlled video generation methods.
For I2V, we compare against CamI2V~\cite{zheng2024cami2v}, MotionClone~\cite{ling2024motionclone}, and CamCloneMaster~\cite{luo2025camclonemaster}. 
CamI2V conditions generation on explicit camera information with geometry-aware attention, while MotionClone is a training-free framework that transfers motion directly from a reference video through temporal attention cues. 
CamCloneMaster also transfers camera motion from a reference video, but does so by jointly processing reference and target tokens in a unified attention framework. For V2V, we compare against DaS~\cite{gu2025diffusion}, TrajectoryCrafter~\cite{yu2025trajectorycrafter}, ReCamMaster~\cite{bai2025recammaster}, and CamCloneMaster~\cite{luo2025camclonemaster}. 
DaS uses 3D tracking videos as motion guidance, TrajectoryCrafter relies on explicit geometry and rendered point-cloud cues, ReCamMaster conditions generation on target camera trajectories together with source video features. 
All methods are evaluated using their official implementations and pretrained models.

\noindent\textbf{Quantitative Comparison.}
As shown in \cref{tab:main_results}, TriMotion achieves the best overall balance between visual quality, temporal stability, and camera-motion accuracy across both I2V and V2V settings. 
The pose-conditioned variant performs best on most metrics, while the video- and text-conditioned variants remain competitive, suggesting that the shared motion space provides an effective control signal across all three modalities. 
We attribute this advantage to the unified motion embedding space and the latent motion consistency loss, which together reduce motion drift during training. 
While TrajectoryCrafter achieves competitive camera accuracy via explicit 3D point-cloud guidance, such rendering often introduces occlusion artifacts and texture degradation, undermining perceptual quality and temporal stability under challenging viewpoint changes.

\begin{figure}[t]
\centering
\includegraphics[width=\columnwidth]{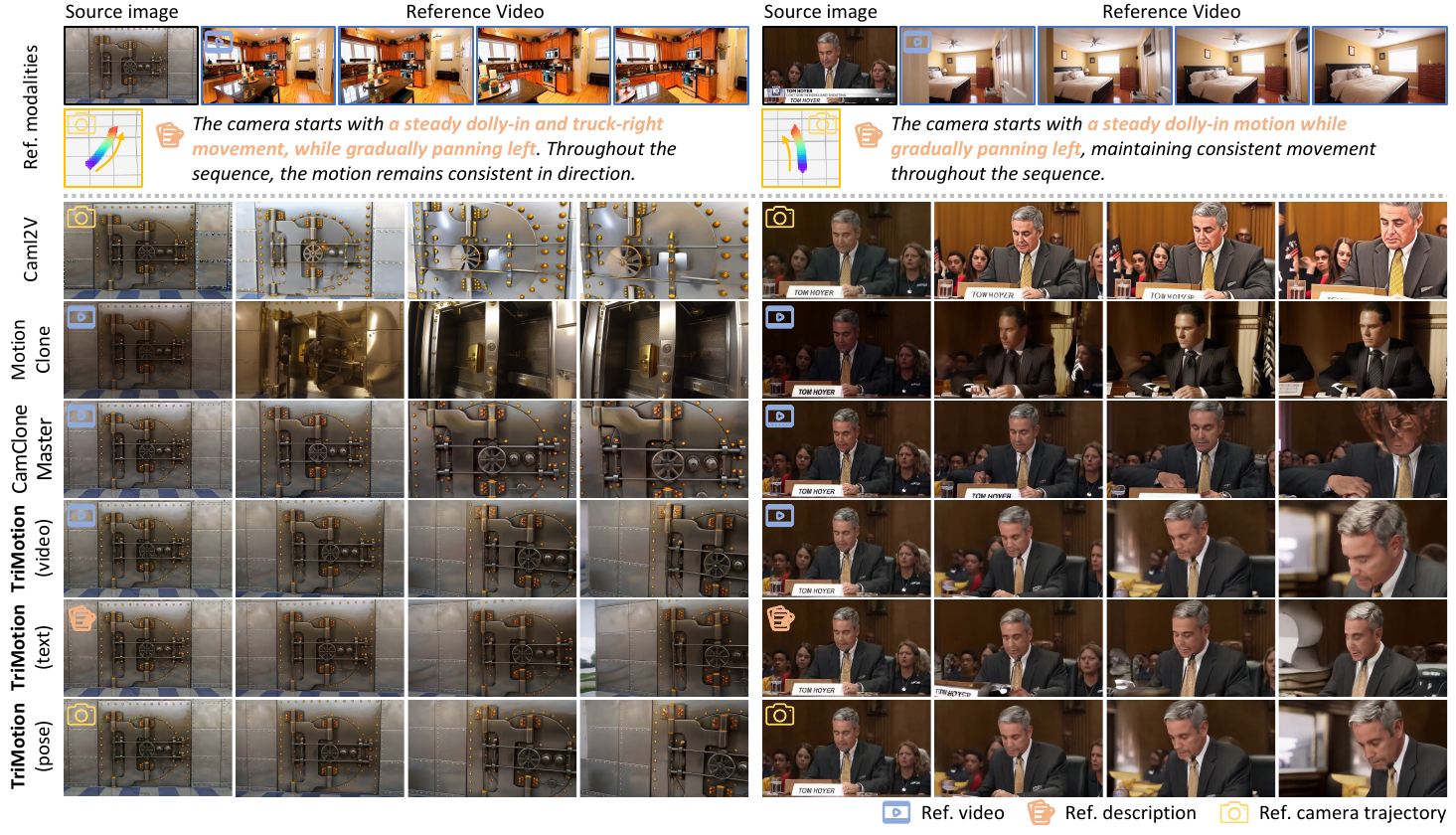}
\vspace{-6mm}
\caption{Qualitative comparison for Camera-controlled I2V generation results.}
\vspace{-4mm}
\label{fig:i2v}
\end{figure}

\begin{figure}[t]
\centering
\includegraphics[width=\columnwidth]{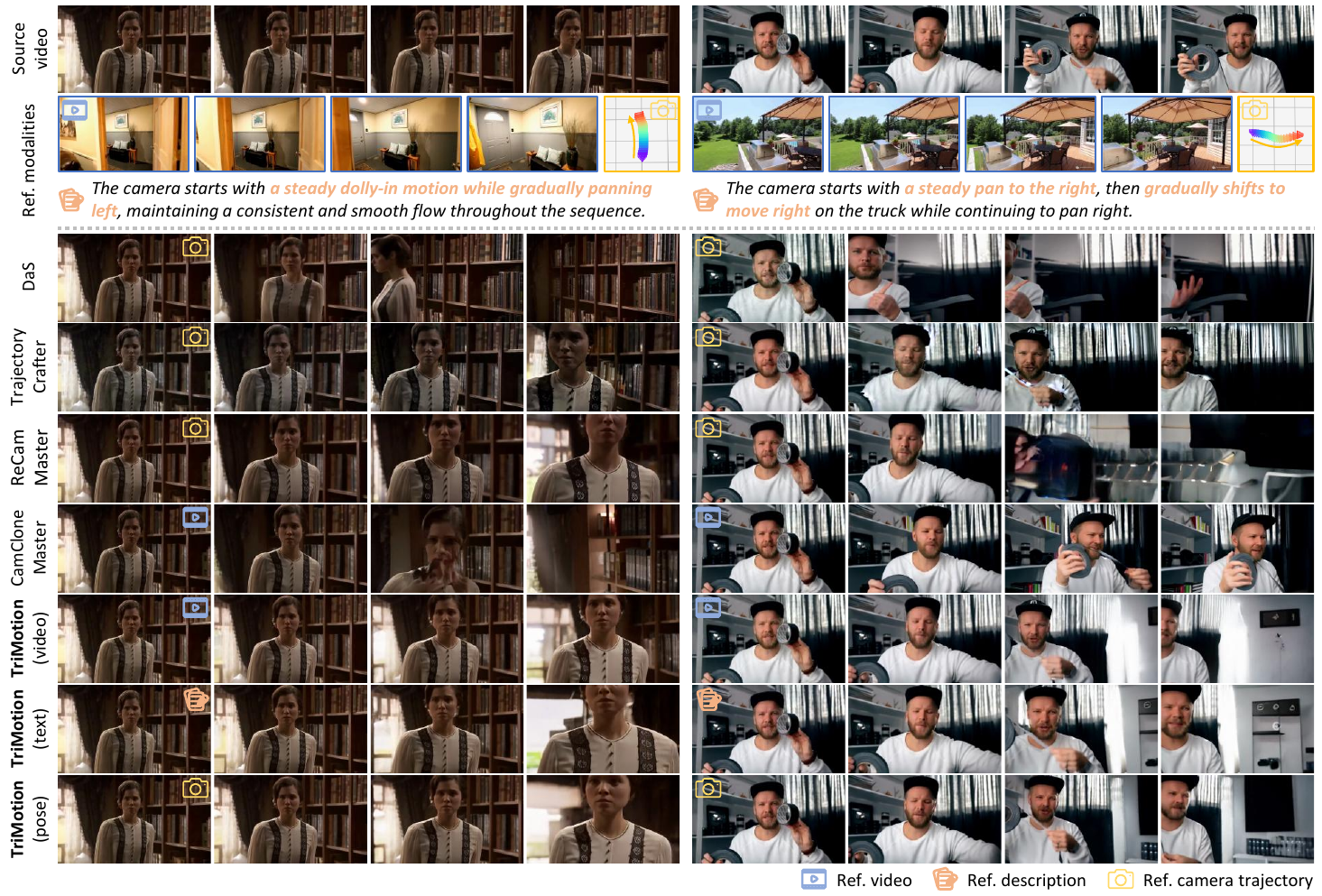}
\vspace{-6mm}
\caption{Qualitative comparison for Camera-controlled V2V generation results.}
\label{fig:v2v}
\end{figure}

\noindent\textbf{Qualitative Comparison.}
Fig.~\ref{fig:i2v} and Fig.~\ref{fig:v2v} present qualitative comparisons on I2V and V2V, respectively. 
Across all input modalities, TriMotion generates cleaner videos with more stable motion and follows the target camera trajectory more faithfully.
CamI2V and ReCamMaster generally follow the target pose reasonably well, but still exhibit visible artifacts in challenging cases. 
MotionClone and CamCloneMaster capture the coarse motion trend, yet often drift from the desired trajectory. 
TrajectoryCrafter achieves strong camera control, but large viewpoint changes reveal noticeable artifacts, while DaS often struggles to preserve the source video content. 
Overall, TriMotion follows compound and long-horizon camera motions more faithfully while maintaining better scene structure and temporal coherence.

\input{Tables/ablation}

\subsection{Ablation Study}
\label{sec:ablation}

To validate our conditioning design and latent motion consistency, we perform ablations in the V2V setting with pose conditioning, since alternative variants require a source motion embedding that is not naturally available in I2V.

As shown in Tab.~\ref{tab:ablation}, concatenation improves several visual metrics but degrades camera-motion accuracy, suggesting that jointly providing source and target embeddings introduces competing motion cues and biases the model toward source preservation. 
By contrast, target-only conditioning provides the cleanest motion signal and the best overall trade-off. 
Using the difference vector is less effective, likely because it removes absolute motion structure and retains only relative displacement.

We further ablate the latent motion consistency loss by training with and without $\mathcal{L}_{\mathrm{motion}}$. 
Removing this loss consistently degrades camera accuracy and slightly weakens visual and temporal quality, indicating that motion conditioning alone is insufficient and that the proposed constraint improves trajectory fidelity by aligning the predicted clean latent with the target motion embedding.

\input{Tables/analysis}
\subsection{Analysis on Unified Motion Embedding Space}
\label{sec:analysis}

We analyze the unified motion space from two perspectives: cross-modal alignment and geometric fidelity. 
For cross-modal alignment, we use the validation split of the Motion Triplet Dataset, where modality pairs from the same trajectory are treated as positives and pairs from different trajectories within the same scene are treated as negatives. 
We report cosine similarity and Recall@1 across modalities. 
As shown in Tab.~\ref{tab:analysis}, the learned embeddings exhibit a clear separation between positive and negative pairs and achieve strong retrieval performance, indicating that they capture motion patterns rather than scene identity.

We further evaluate geometric fidelity through pose regressor in Sec.~\ref{sec:unified_space} on the evaluation set from RealEstate10K. 
The results show that the learned embeddings preserve meaningful geometric information and remain competitive with strong feed-forward pose estimation models, including VGGT~\cite{wang2025vggt} and MegaSaM~\cite{li2025megasam}. 
Notably, even text embeddings can regress camera poses reasonably well, suggesting that the language representation also captures underlying motion geometry.

\begin{figure}[t]
\centering
\includegraphics[width=\columnwidth]{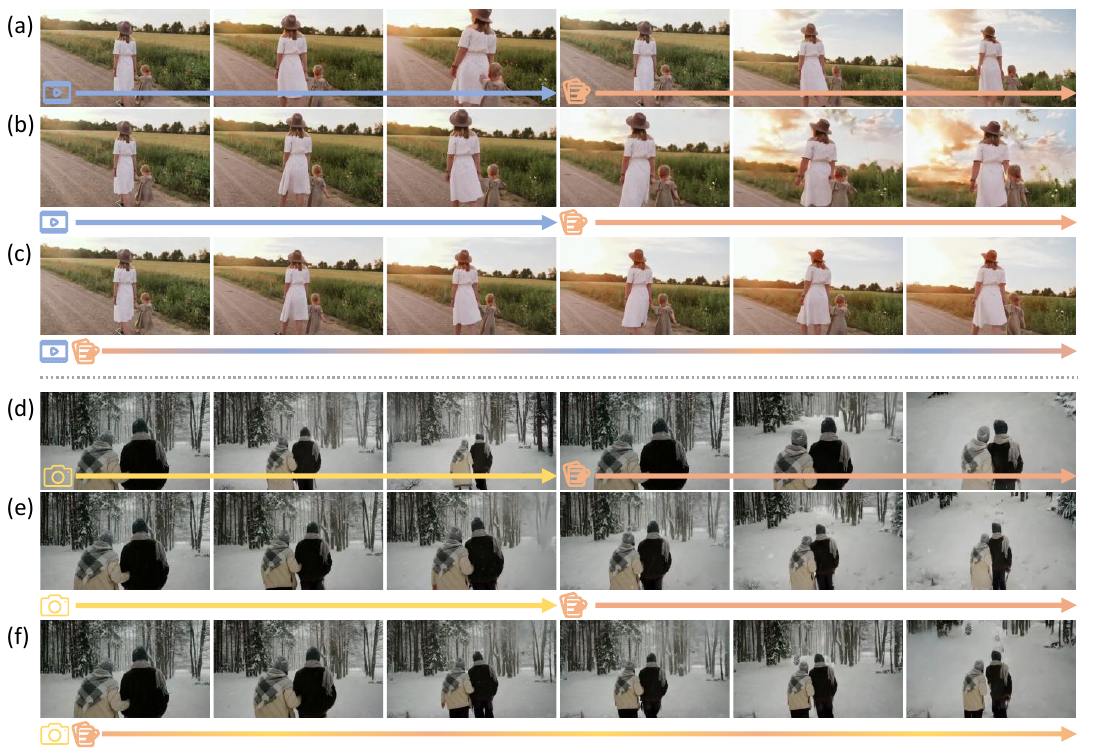}
\vspace{-6mm}
\caption{\textbf{Applications to cross-modal motion composition for camera-controlled V2V generation.}
(a) and (d) show camera-controlled V2V results from a single modality, while (b) and (e) illustrate sequential composition, and (c) and (f) illustrate linear interpolation between motion embeddings.}
\label{fig:application}
\vspace{-5mm}
\end{figure}

\subsection{Applications: Cross-Modal Motion Composition}
\label{sec:application}

A key advantage of the unified motion embedding space is that motion representations from different modalities become directly composable. 
Since video, pose, and text inputs are aligned in the same motion space, their embeddings can be manipulated at inference time to enable flexible camera control beyond single-modality conditioning. 

Given a pair of motion embeddings $\mathbf{e}_a$ and $\mathbf{e}_b$ from different modalities, we explore two composition settings. 
First, we study sequential motion composition by concatenating the two motion sequences, where the second motion is offset by the final state of the first to ensure continuity. 
Second, we study cross-modal motion interpolation by linearly interpolating between $\mathbf{e}_a$ and $\mathbf{e}_b$, which produces camera motions that smoothly blend the two inputs. 
As shown in Fig.~\ref{fig:application}, the unified motion space supports both coherent multi-stage trajectories and smooth interpolation without retraining or manual trajectory design.

\input{Tables/userstudy}

\subsection{User Study}
\label{sec:user_study}

We conduct a user study on Amazon Mechanical Turk (MTurk)~\cite{crowston2012amazon} with 20 participants to evaluate three aspects of our framework: 
(1) the quality of the motion descriptions in the Motion Triplet Dataset, 
(2) the effectiveness of cross-modal motion composition, and
(3) human preference on camera-motion following and perceptual quality.
All evaluation items are randomly drawn from the corresponding set and presented in randomized order.

\noindent\textbf{Motion-description quality.}
We evaluate 30 video-text pairs from the Motion Triplet Dataset, considering the short and detailed descriptions separately. 
For each pair, participants rate how well the text matches the camera motion in the video on a five point scale, where 1 and 5 indicate a poor match and an excellent match, respectively. 
The results in Tab.~\ref{tab:user_study} indicate that the proposed geometry-grounded captioning pipeline reliably captures the underlying camera trajectories.

\noindent\textbf{Cross-modal motion composition.}
We further assess the composability of the unified motion space using 10 sequential-composition examples and 10 interpolation examples. 
For each result, users rate on a five-point scale whether the generated video clearly executes the two motions in sequence in the sequential-composition setting or reflects both input motions in the interpolation setting. 
As shown in Tab.~\ref{tab:user_study}, the learned motion space supports flexible cross-modal motion composition.

\noindent\textbf{Human evaluation of generated videos.}
For generation quality, we consider 10 examples from each of the I2V and V2V settings. 
Participants are first shown a reference clip illustrating the target camera motion and are then presented with videos generated by TriMotion and the baseline methods similar to previous works~\cite{park2024kinetic, lee2026universal, jeon2026rebalancing}. 
They complete two evaluation tasks independently by selecting all videos that 
(1) contain noticeable visual artifacts and 
(2) faithfully follow the target camera motion. 
We report the resulting selection rates in Tab.~\ref{tab:user_study}, where lower is better for artifact selection and higher is better for motion-following selection. 
The results show that TriMotion generates high-quality videos while following the target camera motion more faithfully than prior methods.

\section{Conclusion}

We present \textit{TriMotion}, a unified framework for cross-modal camera motion control in video generation. 
By aligning video, pose, and text in a shared motion embedding space, TriMotion enables flexible and consistent camera control across heterogeneous inputs. 
We further introduce a latent motion consistency constraint that improves trajectory fidelity during generation. 
Experiments on both I2V and V2V show that TriMotion achieves a strong balance among perceptual quality, temporal coherence, and camera-motion accuracy. 
In addition, the shared motion space supports practical applications such as sequential motion composition and cross-modal motion interpolation, highlighting the flexibility of the proposed framework.

\noindent\textbf{Limitations and Future Work.}
Although TriMotion effectively transfers smooth camera motions, extreme trajectories can still produce boundary outpainting artifacts when large unseen regions must be synthesized. 
A promising direction for future work is to integrate our framework with explicit 3D background outpainting or temporal extrapolation modules to improve robustness under large geometric transformations. 
It would also be interesting to extend the framework beyond camera motion and jointly model cinematic attributes~\cite{shin2025video, Shin2024bw} such as color grading, which could enable richer transfer of filmic style and mood together with camera dynamics.

%
%
\newpage
\bibliographystyle{unsrt}
\bibliography{main}
\newpage
\input{appendix}

\end{document}

%% file: Tables/quantitative.tex
\begin{table*}[t]
\caption{Quantitative results for I2V $\&$ V2V tasks. (Best: \textbf{Bold}, Second best: \underline{Underline})}
\vspace{-3mm}
\centering\large
\resizebox{\linewidth}{!}{
\begin{tabular}{cc|cccc| cc|ccc|cc}
\toprule
\multirow{2}{*}{Method\vspace{-15pt}} 
& \multirow{2}{*}{\tworow{Target}{Modality}\vspace{-15pt}~} 
& \multicolumn{4}{c}{Visual Quality} 
& \multicolumn{2}{c}{Dynamic Quality}
& \multicolumn{3}{c}{Motion Accuracy}
& \multicolumn{2}{c}{Source Preservation}
\\ 
\cmidrule{3-6} \cmidrule{7-8} \cmidrule{9-11} \cmidrule{12-13}
&& ~FVD~$\downarrow$~ & ~FID~$\downarrow$~ & ~\tworow{CLIP}{Score}~$\uparrow$~ & ~\tworow{Vbench}{Visual}~$\uparrow$~  
&~CLIP-F~$\uparrow$~& ~\tworow{Vbench}{Dynamic}~$\uparrow$~
&~$E_{\mathrm{rot}}~$~$\downarrow$ & ~$E_{\mathrm{trans}}$~$\downarrow$~ & ~CamMC~$\downarrow$~ 
& ~FVD-V~$\downarrow$~ & ~CLIP-V~$\uparrow$~\\ 
\midrule
&\multicolumn{12}{c}{\textbf{Image-to-Video (I2V)}} \\ 
\midrule
CamI2V~\cite{zheng2024cami2v} & Pose & 361.95 & \underline{39.56} & \underline{29.32} & 0.5655
& 0.9834 & 0.8688 
& 1.1890 & 3.9492 & 4.5648
& - & - \\
MotionClone~\cite{ling2024motionclone} & Video& 601.77 & 50.00 & 28.72 & 0.5826 
& \underline{0.9843} & 0.7887
& 2.1734 & 6.4202 & 7.4974
& - & - \\
CamCloneMaster~\cite{luo2025camclonemaster}& Video & \textbf{312.13 }& 39.68 & 29.00 & 0.5555
& 0.9823 & 0.9042 
& 1.5235 & 4.0527 & 4.9585
& - & - \\
\midrule
TriMotion & Text & 348.63 & 40.23 & 28.41 & \underline{0.5594}
& \textbf{0.9844} & \textbf{0.9249 }
& 1.6875  & 4.0148  & 5.0989
& - & - \\
TriMotion & Video & 337.09 & 39.68 & 29.23 & 0.5591 
& 0.9841 & 0.9128 
& \underline{1.1710}  & \underline{3.8465}  & \underline{4.4117} 
& - & - \\
TriMotion & Pose & \underline{317.85} & \textbf{39.11} & \textbf{29.37} & \textbf{0.5620}
& \underline{0.9843} & \underline{0.9168} 
& \textbf{1.0113} & \textbf{3.8441} & \textbf{4.3087}
& - & - \\

\midrule
&\multicolumn{12}{c}{\textbf{Video-to-Video (V2V)}} \\ 
\midrule
DaS~\cite{gu2025diffusion} & Pose &274.22 & 38.21 & 29.29 & 0.4953
& 0.9744 & 0.8893 
&2.7558 & 7.7572 & 9.2675
&270.65&0.8204\\
TrajectoryCrafter~\cite{yu2025trajectorycrafter}& Pose & 299.96 & 39.51 & \underline{29.40} & 0.5680 
& 0.9764 & 0.9026 
&\textbf{0.6954} & \underline{3.2621} & \textbf{3.5711} 
&267.18&0.8794\\
ReCamMaster~\cite{bai2025recammaster} & Pose & 253.65 & 38.77 & 29.13 & 0.5696
& \underline{0.9883} & 0.9277 
&1.3184& 3.6559 & 4.3611 
&233.58&0.8833\\
CamCloneMaster~\cite{bai2025recammaster} & Video & 241.38 & \underline{38.64} & 29.24 & 0.5725
& 0.9878 &0.9235
&1.5040 &3.8987 & 4.7964 
&195.89&0.8791\\
\midrule

TriMotion & Text & \underline{237.87} & 38.65 & 29.10 & \textbf{0.5820}
& 0.9890 & \underline{0.9345}
&1.6030 & 3.7525 & 4.7778 
&\underline{186.73}&\underline{0.9041} \\

TriMotion & Video & 254.55 & 38.86 & 29.28 & 0.5728
& 0.9882 & 0.9258 
&1.1795 & 3.6468 & 4.2384 
&216.55&0.8851\\
TriMotion & Pose& \textbf{221.59 }& \textbf{38.16 }& \textbf{29.43} & \underline{0.5767}
& \textbf{0.9895} & \textbf{0.9372 }
&\underline{0.9488} & \textbf{3.2356} & \underline{3.6797}
&\textbf{172.25}&\textbf{0.9071}\\
\bottomrule
\end{tabular}}
\vspace{-6mm}
\label{tab:main_results}
\end{table*}

%% file: Tables/ablation.tex
\begin{table*}[t]
\caption{Quantitative results of our ablation study on conditioning design and latent motion consistency. (Best: \textbf{Bold}, Second best: \underline{Underline})}
\vspace{-3mm}
\centering\large
\resizebox{\linewidth}{!}{
\begin{tabular}{c|cccc| cc|ccc|cc}
\toprule
\multirow{2}{*}{Method\vspace{-15pt}} 
& \multicolumn{4}{c}{Visual Quality} 
& \multicolumn{2}{c}{Dynamic Quality}
& \multicolumn{3}{c}{Motion Accuracy}
& \multicolumn{2}{c}{Source Preservation}
\\ 
\cmidrule{2-5} \cmidrule{6-7} \cmidrule{8-10} \cmidrule{11-12}
&~FVD~$\downarrow$~ & ~FID~$\downarrow$~ & ~\tworow{CLIP}{Score}~$\uparrow$~ & ~\tworow{Vbench}{Visual}~$\uparrow$~  
&~CLIP-F~$\uparrow$~& ~\tworow{Vbench}{Dynamic}~$\uparrow$~
&~$E_{\mathrm{rot}}~$~$\downarrow$ & ~$E_{\mathrm{trans}}$~$\downarrow$~ & ~CamMC~$\downarrow$~ 
& ~FVD-V~$\downarrow$~ & ~CLIP-V~$\uparrow$~\\ 
\midrule

Concatenation & \textbf{193.56}& 39.75 & \textbf{29.83} & \textbf{0.5777}
& \textbf{0.9914} & \textbf{0.9461}
&2.1250&10.5327 &11.2641
& \textbf{116.34} & \textbf{0.9214}\\

Subtraction & 265.53 & \textbf{37.12}& \underline{29.80}& 0.5714
&0.9866  & 0.9259
&2.8396 & 9.6904 & 10.9317
&227.69&0.8903\\

w/o $\mathcal{L}_{\textrm{motion}}$& 249.85& 38.48& 29.18 & 0.5683
& 0.9883 & 0.9255
&\underline{1.2307} & \underline{4.2986} & \underline{4.8660}
&214.42 & 0.8828\\

TriMotion & \underline{221.59}& \underline{38.16} & 29.43 & \underline{0.5767}
& \underline{0.9895} & \underline{0.9372}
&\textbf{0.9488} & \textbf{3.2356} & \textbf{3.6797}
&\underline{172.25}&\underline{0.9071}\\

\bottomrule
\end{tabular}}
\vspace{-6mm}
\label{tab:ablation}
\end{table*}

%% file: Tables/analysis.tex
\begin{table*}[t]
\caption{Quantitative results of the analysis on the unified motion embedding space. Positive and Negative refer to the average cosine similarity between positive and negative pairs, respectively. (Best: \textbf{Bold}, Second best: \underline{Underline})}
\vspace{-3mm}
\centering\large
\resizebox{\linewidth}{!}{
\begin{tabular}{c|ccc||c|ccccc}
\toprule
\multirow{2}{*}{Metric\vspace{-8pt}} 
& \multicolumn{3}{c||}{Cross Modal Alignment} 
& \multirow{2}{*}{Metric\vspace{-8pt}} 
& \multicolumn{5}{c}{Geometric Information}
\\ 
\cmidrule{2-4} \cmidrule{6-10} 
&~Video $\leftrightarrow$ Pose~ & ~Video $\leftrightarrow$ Text~ & ~Text $\leftrightarrow$Pose~ 
& & ~VGGT~\cite{wang2025vggt}~ & ~MegaSAM~\cite{li2025megasam}~ & ~~~~Pose~~~~ & ~~~~Video~~~~ & ~~~~Text~~~~\\ 
\midrule

~Positive~ & 0.9605 & 0.8887 & 0.9121 
& ~$E_{\mathrm{rot}}~$~$\downarrow$ &1.6226 & \textbf{0.1478} & \underline{0.4505}& 0.9972 & 1.4300

\\

~Negative~ & 0.0317 & 0.0256 & 0.0243
& ~$E_{\mathrm{trans}}$~$\downarrow$~  & \textbf{0.4459} & 0.7003 &\underline{0.5530}&1.7211&2.9489

\\

~Recall@1~ & 98.50 \% & 98.00 \% & 99.50 \% 
& ~CamMC~$\downarrow$~ &\textbf{0.4966}& 0.7713 &\underline{0.6603}&1.8707&3.8359

\\

\bottomrule
\end{tabular}}
\vspace{-3mm}
\label{tab:analysis}
\end{table*}

%% file: Tables/userstudy.tex
\begin{table*}[t]
\caption{The result of user study. We report the mean and standard deviation for description quality and cross-modal composition.}
\vspace{-3mm}
\centering\large
\resizebox{\linewidth}{!}{
\begin{tabular}{c|cccccccccccccc}
\toprule

Description Quality & \multicolumn{3}{c}{~Short Description~}& \multicolumn{3}{c||}{~Long Description~} & \multicolumn{2}{c|}{~~Cross-modal~~} & \multicolumn{3}{c}{~~~Sequential~~~} & \multicolumn{3}{c}{Interpolation}  \\
\midrule
Score & \multicolumn{3}{c}{3.94 $\pm$ 0.23} & \multicolumn{3}{c||}{4.11 $\pm$ 0.22}&\multicolumn{2}{c|}{Score} & \multicolumn{3}{c}{3.97 $\pm$ 0.23} &\multicolumn{3}{c}{3.88 $\pm$ 0.27} \\

\midrule
\midrule
I2V & \multicolumn{2}{c}{-} & \multicolumn{2}{c}{~CamI2V~} & \multicolumn{2}{c}{~MotionClone~}  &\multicolumn{2}{c}{\tworow{CamClone}{Master}} &\multicolumn{2}{c}{~~\tworow{TriMotion}{Text}~~}&\multicolumn{2}{c}{~~\tworow{TriMotion}{Video}~~}&\multicolumn{2}{c}{~~\tworow{TriMotion}{Pose}}\\
\midrule
Artifact~$\downarrow$ &\multicolumn{2}{c}{-} & \multicolumn{2}{c}{26.0\%} & \multicolumn{2}{c}{63.5\%} & \multicolumn{2}{c}{18.5\%}  &\multicolumn{2}{c}{17.5\%} & \multicolumn{2}{c}{\underline{15.5\%}} & \multicolumn{2}{c}{\textbf{12.5\%}} \\
Motion Accuracy~$\uparrow$&\multicolumn{2}{c}{-} & \multicolumn{2}{c}{78.5\%} & \multicolumn{2}{c}{41.5\%} & \multicolumn{2}{c}{67.5\%}& \multicolumn{2}{c}{81.0\%} & \multicolumn{2}{c}{\underline{82.5\%}} & \multicolumn{2}{c}{\textbf{86.5}\%}  \\ 

\midrule
\midrule
V2V & \multicolumn{2}{c}{~~~~~~DaS~~~~~~}&\multicolumn{2}{c}{\tworow{Trajectory}{Crafter}}&\multicolumn{2}{c}{\tworow{ReCam}{Master} }&\multicolumn{2}{c}{\tworow{CamClone}{Master}}&\multicolumn{2}{c}{~\tworow{TriMotion}{Text}~}& \multicolumn{2}{c}{\tworow{TriMotion}{Video}}&\multicolumn{2}{c}{\tworow{TriMotion}{Pose}} \\
\midrule
Artifact~$\downarrow$&\multicolumn{2}{c}{~44.5\%~} & \multicolumn{2}{c}{20.5\%}& \multicolumn{2}{c}{11.0\%} & \multicolumn{2}{c}{21.5\%} & \multicolumn{2}{c}{15.5\%} & \multicolumn{2}{c}{~~\textbf{8.5\%}}& \multicolumn{2}{c}{~~\underline{9.5\%}} \\
Motion Accuracy~$\uparrow$ & \multicolumn{2}{c}{~43.5\%~} & \multicolumn{2}{c}{69.0\%} & \multicolumn{2}{c}{80.0\%} & \multicolumn{2}{c}{61.5\%} & \multicolumn{2}{c}{74.5\%} & \multicolumn{2}{c}{\textbf{89.5\%}} & \multicolumn{2}{c}{\underline{86.5\%}} \\

\bottomrule
\end{tabular}}
\vspace{-6mm}
\label{tab:user_study}
\end{table*}

%% file: appendix.tex
\setcounter{section}{0}
\setcounter{table}{0}
\setcounter{figure}{0}
\setcounter{equation}{0}
\renewcommand{\thesection}{\Alph{section}}
\renewcommand{\thesubsection}{\Alph{section}.\arabic{subsection}}

\renewcommand{\thefigure}{\Roman{figure}}
\renewcommand{\thetable}{\Roman{table}}
\renewcommand{\theequation}{\Roman{equation}}

\section{Additional Details of the Motion Triplet Dataset}
\label{sec:supp_dataset}

\subsection{Pose Preprocessing and Relative Trajectory Construction}

In the Multi-Cam Video Dataset, each camera trajectory contains 81 pose samples. 
We uniformly subsample the sequence with stride 4 to obtain 21 keyframes for motion caption generation.
This reduces redundancy while preserving the overall motion pattern of the trajectory.
Let $\mathbf{T}_t \in \mathbb{R}^{4\times4}$ denote the homogeneous camera-to-world transform at the $t$-th selected keyframe.
We first convert the dataset coordinates to the coordinate system used for motion analysis by reordering the spatial axes and flipping the vertical axis.
This yields a right-down-forward convention, where the $x$-, $y$-, and $z$-axes correspond to truck, pedestal, and dolly motion, respectively.
Translations are further rescaled from centimeters to meters.
All poses are then expressed relative to the first frame:
\begin{equation}
\tilde{\mathbf{T}}_t = \mathbf{T}_1^{-1}\mathbf{T}_t .
\end{equation}
Each trajectory is therefore represented as a sequence of 21 relative transforms. 
We use the upper $3\times4$ block as the pose representation.

\subsection{Symbolic Motion Phase Extraction}

Before text generation, we convert each relative trajectory into a compact symbolic sequence.
For each keyframe, we decompose the relative pose into rotation $\mathbf{R}_t$ and translation $\mathbf{t}_t$.
The rotation matrices are converted to Euler angles in \texttt{xyz} order, and frame-to-frame increments are computed as
\begin{equation}
\Delta \mathbf{t}_t = \mathbf{t}_{t+1} - \mathbf{t}_t, \qquad
\Delta \boldsymbol{\theta}_t =
\mathrm{wrap}\!\left(
\mathrm{Euler}_{xyz}(\mathbf{R}_{t+1}) - \mathrm{Euler}_{xyz}(\mathbf{R}_t)
\right),
\end{equation}
where $\mathrm{wrap}(\cdot)$ maps angular differences to $[-180^\circ, 180^\circ]$.
To suppress numerical jitter, translation changes smaller than $0.02$ m and rotation changes smaller than $0.3^\circ$ are ignored.
At each step, we identify the motion axis with the largest magnitude and treat it as the dominant motion.
Other axes are retained only if their magnitude is within a factor of 5 of the dominant axis. 
This filtering removes weak residual components while preserving compound camera motion.
The remaining components are mapped to canonical camera operations, as summarized in Tab.~\ref{tab:supp_motion_mapping}.
If no component exceeds the thresholds, the step is labeled as \textit{Stationary}.
The final symbolic representation is a temporally ordered list of motion phases indexed by normalized progress.
Tab.~\ref{tab:symbolic_example} shows an example symbolic sequence generated from a camera trajectory.
This representation preserves the temporal structure of the trajectory and provides a compact symbolic description of the motion.

\begin{table}[t]
\centering
\caption{Mapping from relative motion components to canonical camera operations used in the symbolic trajectory representation.}
\label{tab:supp_motion_mapping}
\begin{tabular}{cc}
\toprule
~~~~~~~~Relative motion component~~~~~~~~ & ~~~~~~~~Canonical operation~~~~~~~~ \\
\midrule
$+\Delta x$ / $-\Delta x$ & Truck Right / Truck Left \\
$+\Delta y$ / $-\Delta y$ & Pedestal Down / Pedestal Up \\
$+\Delta z$ / $-\Delta z$ & Dolly In / Dolly Out \\
$+\Delta \theta_x$ / $-\Delta \theta_x$ & Tilt Up / Tilt Down \\
$+\Delta \theta_y$ / $-\Delta \theta_y$ & Pan Right / Pan Left \\
$+\Delta \theta_z$ / $-\Delta \theta_z$ & Roll CW / Roll CCW \\
\bottomrule
\end{tabular}
\end{table}

\begin{table}[t]
\centering
\caption{Example symbolic motion sequence extracted from a camera trajectory.}
\label{tab:symbolic_example}
\begin{tabular}{ccc}
\toprule
Progress & Motion & Magnitude \\
\midrule
0\%   & Stationary & - \\
15\% & Dolly In & $\Delta z = 0.21m$ \\
35\% & ~~~~~~Dolly In + Pan Right~~~~~~ & ~~~~~~$\Delta z = 0.18m$, $\Delta\theta_y = 3.2^\circ$~~~~~~ \\
60\% & Pan Right & $\Delta\theta_y = 4.1^\circ$ \\
85\% & Dolly Out & $\Delta z = -0.16m$ \\
100\%   & Stationary & -\\
\bottomrule
\end{tabular}
\end{table}

\subsection{Geometry-Grounded Motion Caption Generation}

We translate the symbolic motion sequence into natural language using Qwen3-4B-Instruct~\cite{yang2025qwen3}.
Two prompt templates are used on the same symbolic input.
The first produces a short summary in one or two sentences, and the second produces a detailed description in three to five sentences.
Both prompts explicitly define the coordinate system and instruct the model to describe the temporal evolution of the camera motion rather than numerical values.
In particular, the model is asked to:
(1) distinguish simultaneous motion from sequential motion using connectors such as \textit{while} and \textit{then},
(2) avoid metric quantities such as meters or degrees, and
(3) describe speed changes only qualitatively, for example \textit{accelerating}, \textit{slowing down}, or \textit{moving at a steady pace}.
For the detailed version, we additionally encourage temporal discourse markers such as \textit{Initially}, \textit{Gradually}, and \textit{Towards the end}.
Since the input is symbolic rather than raw pose matrices, the resulting descriptions remain grounded in camera geometry while staying close to realistic user instructions.
We generate one short summary and one detailed description for each camera trajectory.

\subsection{Appearance Caption Generation}

In addition to motion descriptions, we generate appearance captions used as the global appearance description $y$ in Sec.~4.1. 
We use InternVL3-14B~\cite{zhu2025internvl3} as a video captioning model.
For each clip, we uniformly sample 8 frames and resize them to $448 \times 448$.
The ordered frames are presented to the model as a multi-frame input.
The prompt requests a detailed, objective, and visually grounded description of the clip while avoiding narrative fillers.
Specifically, it asks the model to describe:
(1) the environment, including time of day, lighting, weather, and background,
(2) the subjects, including appearance, clothing, and spatial location, and
(3) the actions and events in chronological order.
These captions are used only for appearance conditioning and remain separate from the motion descriptions used for motion alignment.

\section{Additional Implementation Details}
\label{sec:supp_imp}

We provide the architectural configurations of the proposed modules together with the training setup used in our experiments.

\subsection{Unified Motion Space}

Each modality encoder is followed by a temporal Transformer with $2$ layers and $8$ attention heads.
The hidden dimension is $2048$ for the video branch and $768$ for the text and pose branches.
In the video branch, we build on top of the VGGT~\cite{wang2025vggt} feature aggregation module with input image size $518$, patch size $14$, and internal embedding dimension $1024$.
In the text branch, $21$ learnable motion queries cross-attend to contextualized token embeddings from a frozen T5-base encoder~\cite{raffel2020exploring}.
The resulting features are then processed by a $2$-layer Transformer with hidden dimension $768$.
In the pose branch, camera poses are first projected by a frame-wise MLP with two hidden layers of dimension $256$ before temporal aggregation.
To preserve geometric fidelity, we attach a shared pose prediction head to the frame-level embeddings of all modalities.
This head is implemented as a $2$-layer MLP with dimensions $768 \rightarrow 256 \rightarrow 12$, with a ReLU activation and dropout rate $0.1$.
The alignment objective uses $\lambda_t=0.1$, $\lambda_p=1.0$, and temperature $\tau=0.07$.

\subsection{Motion Embedding Predictor}

The motion embedding predictor $\mathcal{M}_{\mathrm{pred}}$ first extracts latent features using two 3D convolution layers with hidden dimension $512$, kernel size $(1,3,3)$, spatial padding $(0,1,1)$, GroupNorm with $8$ groups, and GELU activations.
The resulting features are then processed by spatial and temporal Transformer encoders.
Both Transformers use $2$ layers, $8$ attention heads, feed-forward dimension $2048$, and dropout rate $0.1$.
The final features are projected from dimension $512$ to $768$.
Both global and frame-level embeddings are $L_2$-normalized before the cosine-similarity objective is applied.
For frame-level alignment, we use $\lambda_h = 0.5$.

\subsection{Diffusion Backbone}

We use $\lambda_m=0.1$ for the latent motion consistency loss.
During training, I2V and V2V inputs are sampled with equal probability as described in the main paper. 
The three conditioning modalities are also sampled with equal probability.
At inference time, we use $50$ flow-matching steps with classifier-free guidance scale $5.0$ and sigma shift $5.0$.
All reported results use $81$-frame generation and the same inference hyperparameters across all three conditioning modalities.

\subsection{Reproducibility Notes}

All experiments are conducted on 4 NVIDIA H200 GPUs.
We use AdamW with learning rate $1\times10^{-4}$, bf16 mixed-precision training, and gradient clipping with maximum norm 1.0.
The unified motion space is trained for 100 epochs, the motion embedding predictor for 10 epochs, and the diffusion backbone for 10K iterations.
The unified motion space is trained with a batch size of 24 per GPU.
The motion embedding predictor uses a batch size of 8 per GPU.
For the diffusion backbone, the per-device batch size is 4 with gradient accumulation over 4 steps.
Random seeds are fixed to 42 for all reported experiments.

\section{Additional Experiment Details}
\label{sec:supp_exp}
We summarize the computation protocol of the evaluation metrics used in our experiments.

\subsection{Visual quality}
We report FVD~\cite{unterthiner2018towards}, FID~\cite{heusel2017gans}, CLIP Score~\cite{radford2021learning} and Image Quality and Aesthetic Quality from VBench~\cite{huang2024vbench} for visual quality.
FVD measures the distributional similarity between generated and real videos by computing the Fr\'echet distance between their feature distributions in the I3D~\cite{carreira2017quo} feature space. 
FID measures the distributional discrepancy between generated and real frames by computing the Fr\'echet distance between frame-level feature statistics extracted from an Inception-v3 network~\cite{szegedy2016rethinking}. 
CLIP Score measures semantic alignment between the conditioning text and the generated frames by averaging the cosine similarity between the CLIP text embedding of the appearance description and the CLIP image embeddings of the generated frames. 
Image Quality evaluates low-level perceptual distortions such as blur, noise, and exposure artifacts using the MUSIQ predictor~\cite{ke2021musiq}. 
Aesthetic Quality evaluates the perceptual and artistic quality of generated frames using the LAION aesthetic predictor~\cite{LAION-Aesthetic}.

\subsection{Dynamic quality}
For dynamic quality, we evaluate CLIP-F~\cite{bai2025recammaster} together with four VBench metrics: Motion Smoothness, Temporal Flickering, Subject Consistency, and Background Consistency.
CLIP-F measures short-term temporal consistency by computing the average cosine similarity between CLIP image embeddings of consecutive generated frames. 
Motion Smoothness evaluates whether the generated motion evolves smoothly and plausibly over time using the motion prior of a video frame interpolation model~\cite{li2023amt}. 
Temporal Flickering measures frame-to-frame visual instability by computing the intensity difference between adjacent frames. 
Subject Consistency evaluates whether the appearance of the main subject remains stable across the video using DINO~\cite{caron2021emerging} feature similarity, while Background Consistency measures the temporal stability of background using CLIP feature similarity across frames.

\subsection{Camera Motion Accuracy}
We estimate camera poses of generated videos using MegaSaM~\cite{li2025megasam} and compare with the target trajectory in relative pose space. 
To account for differences in global coordinate frame and scale, both predicted and ground-truth trajectories are converted to relative poses before evaluation. 
We use three metrics to evaluate camera motion accuracy: Rotation Error, Translation Error, and Camera Motion Consistency (CamMC).
Rotation Error ($E_{\mathrm{rot}}$) measures the discrepancy between relative rotations, Translation Error ($E_{\mathrm{trans}}$) measures the $L_2$ difference between relative translations after scale normalization, and CamMC evaluates the overall agreement between predicted and target relative pose vectors following the protocol of prior camera-control work~\cite{zheng2024cami2v}.

\subsection{Source Content Preservation for V2V}
In the V2V setting, we further report FVD-V~\cite{xie2024sv4d} and CLIP-V~\cite{bai2025recammaster} for measuring source-content preservation.
FVD-V measures FVD between generated videos and corresponding source videos.
CLIP-V measures source-content preservation by averaging the cosine similarity between CLIP image embeddings of frames from the source video and the corresponding frames from the generated video.

\section{Analysis of the Motion Embedding Predictor}
We evaluate the motion embedding predictor by measuring how well its latent-space predictions align with the teacher motion embeddings from the video branch of the unified motion encoder.
Given a video, we first obtain the teacher motion embedding $\mathbf{e}_{\mathrm{video}}$ using the video encoder $\Phi_{\mathrm{video}}$. 
In parallel, we encode the video with the 3D VAE to obtain its latent representation $\mathbf{z}$, and feed $\mathbf{z}$ into the motion embedding predictor $\mathcal{M}_{\mathrm{pred}}$ to produce the predicted embedding sequence $\hat{\mathbf{e}}$. 
We then compute the cosine similarity between the two embeddings at both the global and frame-level representations. 
The evaluation is conducted on the validation split of the Motion Triplet Dataset.

The predictor achieves a global cosine similarity of $0.940$ and an average frame-level cosine similarity of $0.989$. 
The results indicate that the predictor reliably recovers motion information from VAE latents, which supports 
the effectiveness of the proposed latent motion consistency objective during diffusion training.